\documentclass[sigconf]{acmart}

\usepackage{booktabs} % For formal tables
\usepackage{algorithmic}
\usepackage{algorithm}
\usepackage{graphicx}
\usepackage{subfigure}

\copyrightyear{2019}
\acmYear{2019} 
\acmConference[WWW '19]{Proceedings of the 2019 World Wide Web Conference}{May 13--17, 2019}{San Francisco, CA, USA}
\acmPrice{}
\acmDOI{10.1145/3308558.3313586}
\acmISBN{978-1-4503-6674-8/19/05}

\usepackage{balance}

\begin{document}

\title{Triple Trustworthiness Measurement for Knowledge Graph}

\author{Shengbin Jia}
\authornote{This author is the one who did all the really hard work.}
\affiliation{%
	\institution{Tongji University}
	\city{Shanghai}
	\country{China}
}
\email{shengbinjia@tongji.edu.cn}

\author{Yang Xiang}
\affiliation{%
	\institution{Tongji University}
	\city{Shanghai}
	\country{China}
}
\email{shxiangyang@tongji.edu.cn}

\author{Xiaojun Chen}
\affiliation{%
	\institution{Tongji University}
	\city{Shanghai}
	\country{China}
}
\email{xiaojunchen@tongji.edu.cn}

\begin{abstract}
The Knowledge graph (KG) uses the triples to describe the facts in the real world. It has been widely used in intelligent analysis and applications. However, possible noises and conflicts are inevitably introduced in the process of constructing. And the KG based tasks or applications assume that the knowledge in the KG is completely correct and inevitably bring about potential deviations. In this paper, we establish a knowledge graph triple trustworthiness measurement model that quantify their semantic correctness and the true degree of the facts expressed. The model is a crisscrossing neural network structure. It synthesizes the internal semantic information in the triples and the global inference information of the KG to achieve the trustworthiness measurement and fusion in the three levels of entity level, relationship level, and KG global level. We analyzed the validity of the model output confidence values, and conducted experiments in the real-world dataset FB15K (from Freebase) for the knowledge graph error detection task. The experimental results showed that compared with other models, our model achieved significant and consistent improvements.
\end{abstract}




\maketitle

\section{Introduction}

A knowledge graph (KG), aims to describe the various entities or concepts and their relationships existing in the objective world~\cite{Xu2016a}, which lays the foundation for the knowledge-based organization and intelligent application in the Internet age with its powerful semantic processing capabilities and open organization capabilities. It has attracted increasing attention in academic and industrial circles. It usually stores knowledge in the form of triples (head entity, relationship, tail entity), which can be simplified to $(h, r, t)$.

The construction of the preliminary KG has mainly relied on manual annotation or expert supervision~\cite{Qiao2016,Dong2014}. This way is extremely labor-intensive and time-consuming, and can no longer meet the speed of updating and growth of the real-world knowledge~\cite{Dong2014}. Therefore, an increasing number of researchers are committed to productively extracting information directly from unstructured web text, such as ORE~\cite{Banko2007,Fader2011a,Jia2018}, NELL~\cite{Carlson2010}, and to automatically constructing large-scale knowledge graphs, such as Freebase~\cite{Bollacker2007}, DBpedia~\cite{Auer2007}, and Wikidata\footnote{https://www.wikidata.org/wiki/Wikidata}.
However, some noises and errors are inevitably introduced in the process of automation. References~\cite{Liang2017} and~\cite{Heindorf2016Vandalism} verify the existence and problems of errors in the KG. Existing knowledge-driven learning tasks or applications~\cite{Guan2018,Li2016,Lukovnikov2017Neural,Han2018Knowledge}, assume knowledge in the existing KG is completely correct and therefore bring about potential errors~\cite{Xie2017a,Manago1987}.

For a piece of knowledge in a KG, especially from a professional field, it is difficult to clearly determine whether it is true when it is not tested in practice or is not strictly and mathematically proven. Therefore, we introduce the concept of KG triple trustworthiness, which indicates the degree of certainty that the knowledge expressed by the triple is true. It's value is set to be within the interval [0, 1]. The closer the value is to 0, the greater the probability that the triple is in error. Based on this, we can find possible errors in the existing KG and improve knowledge quality in the KG. 

There are intricate relationships among the entities in the KG, the same relationship can occur between different entities, and multiple relationships can associate with the same entity at the same time.
It is a challenge to study how to use appropriate methods to evaluate the trustworthiness for a knowledge triple. We propose a knowledge graph triple trustworthiness measurement model (KGTtm), which is a crisscrossed neural network-based structure. We measure the trustworthy probability from multiple levels, including the entity level (correlation strength between an entity pair), the relationship level (translation invariance of relation vectors), and the KG global level (inference proof of triple related reachable paths). Corresponding to different levels, we generate three essential questions and focus on solving them by designing three kinds of Estimators. Next, a comprehensive triple confidence value is output through a Fusioner. 

The main contributions of this work include: (1) We propose a knowledge graph triple trustworthiness measurement method that makes comprehensive use of the triple semantic information and globally inferring information. We can achieve three levels of measurement and an integration of confidence value at the entity level, relationship level, and the knowledge graph global level.
(2) We empirically verify the effectiveness of the triple trustworthiness on benchmark data created from a real-world, large-scale KG Freebase. Experimental results show that the error or noise instances are assigned low confidence values, meanwhile, high trustworthiness for true triples.
(3) The trustworthiness calculated by the KGTtm could be utilized in knowledge graph construction or improvement. We evaluate the model on the knowledge graph error detection task. Experimental results show that the KGTtm can effectively detect the error triples and achieve promising performances.

\begin{figure}
	\centering
	\includegraphics[width=0.42\textwidth]{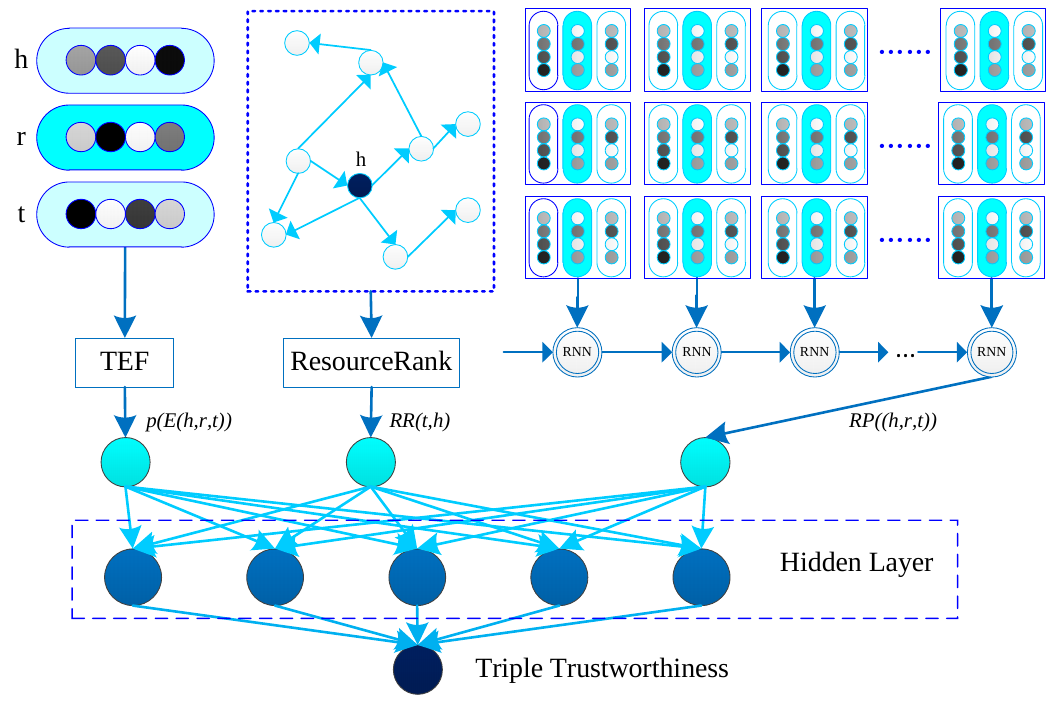}
	\caption{The triple trustworthiness measurement model for KG.\label{fig1}}
\end{figure}

\section{Related Work}
\label{sec:relatedwork}
The concept of Trustworthiness has been applied to knowledge graph related tasks to some extent. Reference~\cite{Xie2017a} proposed a triple confidence awareness knowledge representation learning framework, which improved the knowledge representation effect. There were three kinds of triple credibility calculation methods using the internal structure information of the KG. This method used only the information provided by the relationship, ignoring the related entities. The NELL~\cite{Carlson2010} constantly iterated the extracting template and kept learning new knowledge. It used heuristics to assign confidence values to candidate relations and continuously updated the values through the process of learning. This method was relatively simple but lacked semantic considerations. Dong et al.~\cite{Dong2014} constructed a probabilistic knowledge base (Knowledge Vault), where the reliable probability of a triple was a fusion of some estimators. Several extractors provided a reliability value; meanwhile, a probability could be computed by many prior models which were fitted with existing knowledge repositories in Freebase. This method was tailored for their knowledge base construction and did not have good generalization capabilities. Li et al.~\cite{Li2016} used the neural network method to embed the words in ConceptNet and provide confidence scores to unseen tuples to complete the knowledge base. This method considered only the triples themselves, ignoring the global information provided by the knowledge base.
The above models used the trustworthiness to solve various specific tasks. It shows that the triple trustworthiness is important for applications and research. However, there is a lack of systematic research on the knowledge triple trustworthiness calculation method at present. Our work is devoted to this basic research and proposes a unified measurement model that could facilitate a variety of tasks.

In this work, we verify the effect of the triple trustworthiness on the knowledge graph error detection task. The Knowledge graph error detection task is dedicated to identifying whether a triple in the KG is in error. The existence of noise and errors in the KG is unavoidable. Therefore, error detection is especially important for KG construction and application. The error detection can actually be regarded as a special case of the trustworthiness measurement, which is divided into two kinds of Boolean value types: ``true (trusted)" and ``error (untrusted)". Traditional methods~\cite{Carlson2010,Hoffart2013,Heindorf2016Vandalism,Lehmann2015DBpedia} were still based on manual detection, and the cost was considerable. Recently, some people have begun to study automatic KG error detection methods~\cite{Dong2014,Nickel2015,Liang2017,Shi2016Discriminative}. 
In particular, embedding-based methods~\cite{Li2016,Bordes2013,Toutanova2016} have gained a significant amount of attention. we can efficiently measure the semantic correlations of entities and relations in the vector space. Whether two entities have a potential relationship could be predicted by matrix-vector operations of their corresponding embeddings. They have good efficiency and prospect.
Furthermore, the Knowledge representation learning (KRL) technology is used to project the entities and relations in the KG into a dense, real-valued and low-dimensional semantic embeddings. The main methods include TransE~\cite{Bordes2013}, TransH~\cite{Wang2014}, TransR~\cite{Lin2015}, TransD~\cite{Ji2015}, PTransE~\cite{Lin2015b}, ComplEx~\cite{Trouillon2017} and others.

\section{The Triple Trustworthiness Measurement Model}

The triple trustworthiness measurement (KGTtm) Model for knowledge graph is presented based on a crisscrossing neural network structure, as shown in figure 1. Longitudinally, it can be divided into two levels. The upper is a pool of multiple trustworthiness estimate cells (Estimator). The output of these Estimators forms the input of the lower-level fusion device (Fusioner). The Fusioner is a Multi-layer perceptron to generate the final trustworthiness value for each triple. Viewed laterally, for a given triplet $(h, r, t)$, we consider the triple trustworthiness from three progressive levels and correspondingly answer three hierarchical questions. 1) Is there a possible relationship between entity pairs $(h, t)$? 2) Can a certain relationship r occur between entity pairs $(h, t)$? 3) From a global perspective, can other relevant triples in the KG infer that the triple is trustworthy? For answering these questions we designed three kinds of Estimators, as described below.

\begin{figure*}
	\begin{minipage}{0.3\linewidth}
		\centering
		\includegraphics[width=0.6\textwidth]{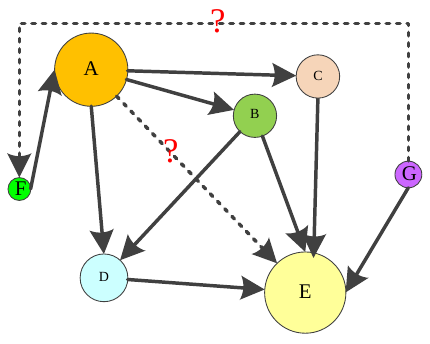}
		\caption{The graph of resource allocation in the ResourceRank algorithm.}
		\label{fig2}
	\end{minipage}%
	\hspace{0.02\linewidth}
	\begin{minipage}{0.3\linewidth}
		\centering
		\includegraphics[width=0.6\textwidth]{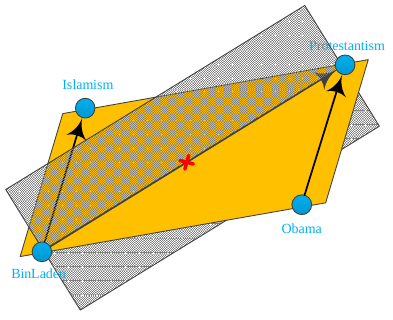}
		\caption{Effects display of the Translation based energy function.}
		\label{fig3}
	\end{minipage}
	\hspace{0.02\linewidth}
	\begin{minipage}{0.3\linewidth}
		\centering
		\includegraphics[width=0.78\textwidth]{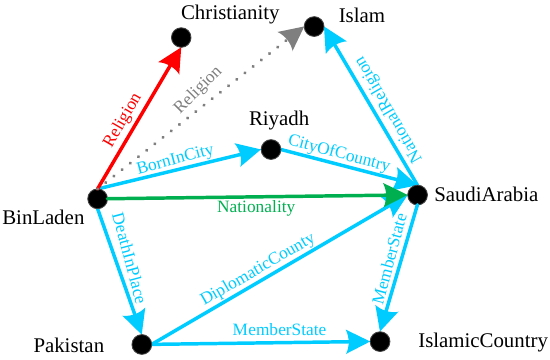}
		\caption{The inference instances for triple trustworthiness.}
		\label{fig4}
	\end{minipage}
\end{figure*}

\subsection{Is there a possible relationship between the  entity pairs?}

We use the association strength between a given entity pair $(h, t)$ to measure the likelihood of an undetermined relationship occurring between the pair. If a pair of entities has heavily weak relevance, it seems to be hopeless that there is a relationship between the entity pairs. The trustworthiness of the triples formed by the entity pair will be greatly compromised. As shown in figure~\ref{fig2}, there are dense edges (relationships) from node (entity) $A$ to node $E$, that is, there is a high association strength about $(A, E)$. We can easily guess that there is a relationship between entity pair $(A, E)$. However, it is impossible to reach $F$ from $G$ following the directed edges. We can also guess that there is no a relationship between $(G, F)$.

%\begin{figure}
%	\includegraphics[width=0.2\textwidth]{figure/jia2.pdf}
%	\caption{The graph of resource allocation in the ResourceRank algorithm.}
%	\label{fig2}
%\end{figure}

We propose an algorithm named \textbf{ResourceRank}, to characterize the association strength between an entity pair according to the idea of Resource allocation~\cite{Xie2017a,Lin2015b,Zhou2007,lu2011link}. The algorithm assumes that the association between entity pairs $(h, t)$ will be stronger, and more resource is passed from the head $h$ through all associated paths to the tail $t$ in a graph. The amount of resource aggregated into $t$ ingeniously indicates the association strength from $h$ to $t$. 

The ResourceRank algorithm mainly includes three steps: 1) Constructing a directed graph centered on the head entity $h$. 2) Iterating the flow of resources in the graph until it converges and calculates the resource retention value of the tail entity $t$. 3) Synthesizing other features and output the likelihood of $(h, ?, t)$.

Specific details are described below:
Each entity is abstracted into a node. If there is a relationship from the entities $e_{1}$ to $e_{2}$, a directed edge will exist from node $e_{1}$ to $e_{2}$. Therefore, the KG can be mapped as a directed graph. This graph is weakly connected, but starting from $h$ each node in the graph can be reached. In the initial state, the resource amount of $h$ is 1, others is 0, and the sum of all nodes is always 1. If a node does not exist in the graph, then its resource is always 0. Moreover, there may be multiple relations between an entity pair $(e_{1}, e_{2})$ but only one directed edge from $e_{1}$ to $e_{2}$ in the graph. Depending on the number of these relations, each edge will have a different bandwidth. The larger the bandwidth is, the more resource flows through the edge.

The resource owned by node $h$ will flow through all associated paths to each other nodes in the entire graph. 
We simulate the flow of resource flowing until distribution steady based on the PageRank~\cite{Page1998,Broder2000} algorithm. The value of the resource on the tail entity is $R\left ( t\mid h \right )$, it is calculated as follows:
\begin{equation}
	R\left ( t\mid h \right )=(1-\theta) \sum_{e_{i}\in M_{t}}\frac{R\left ( e_{i}\mid h \right ) \cdot BW_{e_{i}t}}{OD\left (  e_{i}\right )}+\frac{\theta }{N}.\label{eq1}
\end{equation}
Where, $M_{t}$ is the set of all nodes that have outgoing links to the node $t$, $OD\left (e_{i}\right )$ is the out-degree of the node $e_{i}$  and the $BW_{e_{i}t}$ is the bandwidth from the $e_{i}$ to $t$. Thus, for each node $e_{i}$ in $M_{t}$, the resource transferred from $e_{i}$ to $t$ should be $\frac{R\left ( e_{i}\mid h \right ) \cdot BW_{e_{i}t}}{OD\left (  e_{i}\right )}$. Because the KG noises, the graph is imperfect. And there may be closed loops affecting the resources flowing. In order to improve the model fault-tolerance, we assume that the resource flow from each node may directly jump to a random node with the same probability $\theta$. This part of resource that flows to t randomly is $\frac{1}{N}$, and $N$ is the total number of nodes.

Different states of nodes in the graph reflect information of entity. Considering the following six characteristics: 1) $R\left ( t\mid h \right )$; 2) In-degree of head node $ID (h)$; 3) Out-degree of head node $OD (h)$; 4) In-degree of tail node $ID (t)$; 5) Out-degree of tail node $OD (t)$; 6)The depth from head node to tail node $Dep$, we can construct a feature vector $V$. After being activating, the vector is transformed into a probability value as $RR\left ( h,t \right )$, indicating the likelihood that there may be one or more relationships between the head entity $h$ and the tail entity $t$. This transformation is:
\begin{equation}
\left\{\begin{matrix}
u=\alpha \left ( W_{1}V+b_{1} \right )\\
RR\left ( h,t \right )=W_{2}u+b_{2}\\
\end{matrix}\right.
\end{equation}
Here, $\alpha$  is a nonlinear activation function,  $W_{i}$ and $b_{i}$  are parameter matrices that can be trained during model training. The $RR\left ( h,t \right )$ is within the range [0, 1]. The closer it is to 1, the more likely it is that there is a relationship between $h$ and $t$, which answer the questions shown in the title in the entity layer.

\subsection{Can the determined relationship $r$ occur between the entity pair $(h, t)$ ?}

The above Estimator can only measure the likelihood that one undetermined relationship occurring between the entity pair, but not what kind of relationship. We next calculate the possibility of such a relation $r$ occurring between the entity pair $(h, t)$ by the \textbf{Translation-based energy function (TEF)} algorithm.

Inspired by the translation invariance phenomenon in the word embedding space~\cite{mikolov2013distributed,Mikolov2013}, the relationship in the KG is regarded as a certain translation between entities; that is, the relational vector $\mathbf{r}$ is as the translating operations between the head entity embedding $\mathbf{h}$ and the tail entity embedding $\mathbf{t}$~\cite{Bordes2013}. As illustrated in figure~\ref{fig3}, in the vector space, the same relational vector can be mapped to the same plane and freely translated in the plane to remain unchanged. The triples (BinLaden, Religion, Islam) and (Obama, Religion, Protestantism) should be all correct. However, according to translational invariance of relation vectors, (BinLaden, Religion, Protestantism) must be wrong. Therefore, An trustworthy triple $(h, r, t)$, should satisfy $\mathbf{h}+\mathbf{r} \approx \mathbf{t}$. The energy function is defined as $E(h, r, t) = \left \|\mathbf{h} + \mathbf{r} - \mathbf{t}  \right \|$. The higher the degree of fit between $h$, $r$, and $t$, the smaller the value of $E(h, r, t)$ will be. We believe that the smaller the $E(h, r, t)$ value is, the probability that the relationship $r$ is established between the entity pair $(h, t)$ will be greater, and the trustworthiness of $(h, r, t)$ will be better, and vice versa.

%\begin{figure}
%	\centering
%	\includegraphics[width=0.2\textwidth]{figure/jia3}
%	\caption{Effects display of the Translation-based energy function.}
%	\label{fig3}
%\end{figure}

The TEF algorithm operates as follows: 
Firstly, knowledge representation learning technology is used to implement a low-dimensional distributed representation for entities or relations, and we compute $E(h, r, t)$ for each triple. Then, a modified sigmoid function is used to convert $E(h, r, t)$ into the probability that the entity pair $(h, t)$ constitutes the relationship $r$. The conversion formula is as follows:
\begin{equation}
P(E(h, r, t))=\frac{1}{1+e^{-\lambda \left ( \delta _{r} -E(h, r, t)\right )}}
\end{equation}
Here, $\delta _{r}$  is a threshold related to the relationship $r$. When $E(h, r, t)$ = $\delta _{r}$, the probability value $P$ is 0.5. If $E(h, r, t)$ < $\delta _{r}$, then $P$ > 0.5. The $\lambda$  is a hyperparameter used for smoothing and can be adjusted dynamically along with the model training. The $P(E(h, r, t))$ answers the second question in the relation layer.

\subsection{Can the relevant triples in the KG infer that the triple is trustworthy?}

Inspired by ``social identity'' theory~\cite{turner1986significance,james2015despite}, we make an metaphor: regarding the KG as a social group, where each triple is an individual. The degree of acknowledgements from other individuals to the targeted individual (target triple) reflects whether the targeted individual can properly integrate into the society (i.e., the KG). We believe that only a true triple can achieve popular recognition. Vice versa, if a triple is well accepted, we tend to believe that it is trustworthiness. Therefore, the answer is yes to the question in the title. How to infer the credibility of the target triple by evaluating the acknowledgements of the relevant triples in the KG?

We design a \textbf{Reachable paths inference (RPI)} algorithm to meet it.
There are many substantial multi-step paths from head entities to tail entities, which indicate the semantic relevance and the complex inference patterns among triples~\cite{Pirr2015Explaining}. These reachable paths will be important evidences for judging the triple trustworthiness. For example, as shown in figure~\ref{fig4}, there are multiple reachable paths between entity pairs ``Bin Laden" and ``Saudi Arabia''. According to the path ``Bin Laden $\overset{BornInCity}{\rightarrow}$ Riyadh $\overset{CityOfCountry}{\rightarrow}$ Saudi Arabia", we can firmly infer the fact triple (Bin Laden, Nationality, Saudi Arabia). In addition, we suppose there is a pseudo-triple (Bin Laden, Religion, Christianity) in the KG. The related paths will be very few and illogical, and we should doubt the credibility of this tuple. In contrast, we can find the correct triple (Bin Laden, Religion, Islam) because it gets good acknowledgements depending on multiple reachable paths.
To exploit the reachable paths for inferring triple trustworthiness, we need to address two key challenges:

%\begin{figure}
%	\centering
%	\includegraphics[width=0.2\textwidth]{figure/jia4}
%	\caption{The inference instances for triple trustworthiness.}
%	\label{fig4}
%\end{figure}

\subsubsection{Reachable Paths Selection}

In a large-scale KG, the number of reachable paths associated with a triple may be enormous. It is consuming to weigh all the paths. meanwhile, not all paths are meaningful and reliable. For example, the path ``Bin Laden $\overset{DeathInPlace}{\rightarrow}$ Pakistan $\overset{DiplomaticCountry}{\rightarrow}$ Saudi Arabia'' provided only scarce evidence to reason about the credibility of the triple (Bin Laden, Nationality, Saudi Arabia). Therefore, it is necessary to choose the most efficient reachable paths to use. Previous works believed that the paths that led to lots of possible tail entities were mostly unreliable for the entity pair. They proposed a path-constraint resource allocation algorithm to select relation paths~\cite{Xie2017a,Lin2015b}. Such a method ignored the semantic information of the paths. However, we find that the reliability of the reachable path is actually a consideration of the semantic relevance of the path with the target triple. Therefore, we propose a \textbf{Semantic distance-based path selection} algorithm, which is described as Algorithm~\ref{alg:A}.

\subsubsection{Reachable Paths Representation}

After the paths are selected, it is necessary to map each path to a low-dimensional vector for subsequent calculations. The previous methods~\cite{Xie2017a,Lin2015b} merely considered the relations in the paths. Here, we consider whole triple in the paths, including not only relations but also the head, tail entities, since the entities can also provide significant semantic information. The embeddings of the three elements of each triple are concatenated as a unit $s$. Therefore, a path is transformed into an ordered sequence $S$ = $\begin{Bmatrix}s_{1}, s_{2}, ..., s_{n}\end{Bmatrix}$. 
We use the recurrent neural networks (RNNs)~\cite{hochreiter1997long}, which are good at capturing temporal semantics of a sequence, to learn the semantic information contained in the path.
The RNN layer encodes $s_{t}$  by considering forward information from $s_{1}$ to $s_{t}$. We use the output vector $\mathbf{h_{t}}$ of the last time to represent the semantic information of each path.
We stitch the output $\mathbf{h_{t}}$ of the $TopK$ paths together to form a vector. The vector is nonlinear transformed (using the method in eq(2).) into a recognition value as $RP ((h,r, t))$ , indicating the recognition degree that the relevant triples in the KG infer the target triple being credible.

\begin{algorithm}
	\caption{Reachable Paths Selecting Algorithm}
	\label{alg:A}
	\begin{algorithmic}[1] %这个1 表示每一行都显示数字
		\REQUIRE ~~\\ %算法的输入参数：Input
		
		The knowledge graph (KG); A given target triple $(h, r, t)$.
		
		\ENSURE ~~\\ %算法的输出：Output
		Multiple reachable paths most relevant to target triple.
		
		\STATE Search the reachable paths from $h$ to $t$ in the KG and store in $P_{(h, r, t)}=\begin{Bmatrix}p_{1}, ... , p_{n} \end{Bmatrix}$; 
		\label{ code:fram:search }
		
		\STATE For each $p_{i}=\begin{Bmatrix}(h, l_{1}, e_{1}), (e_{1}, l_{2}, e_{2}), ... , (e_{n-1}, l_{n}, t)\end{Bmatrix}$, calculate\\
		\hangafter=1
		1) the semantic distance between the $r$ and all relations in $p_{i}$, as, 
		{\centering $SD(p_{i}(l),r)=\frac{1}{n}\sum_{l_{j}\in p_{i}(l)}\frac{r\cdot l_{j}}{\left \| r \right \|\left \| l_{j} \right \|}$;\\}
		\hangafter=1
		2) the semantic distance between the $t$ and all head entities in $p_{i}$, as, 
		{\centering $SD(p_{i}(e),t)=\frac{1}{n}\sum_{e_{j}\in p_{i}(e)}\frac{t\cdot e_{j}}{\left \| t \right \|\left \| e_{j} \right \|}$;\\}
		\hangafter=1
		3) the semantic distance between the $h$ and all tail entities in $p_{i}$, as, 
		{\centering $SD(p_{i}(e),h)=\frac{1}{n}\sum_{e_{j}\in p_{i}(e)}\frac{h\cdot e_{j}}{\left \| h \right \|\left \| e_{j} \right \|}$;\\}
		\label{code:fram:for}
		\STATE Calculate the average distance\\
		{\centering $\bar{SD}(p_{i}) = \frac{1}{3}(SD(p_{i}(e), t) + SD(p_{i}(l), r) +SD(p_{i}(e), h))$; \\}
		\label{code:fram:avg}
		\STATE Select first $TopK$ paths with the highest $\bar{SD}(p_{i})$ scores.
		
		\label{code:fram:select}
		
		\STATE Return $\begin{Bmatrix}p_{i}\mid 1\leqslant i \leqslant TopK, Sort(\bar{SD}(p_{i}), descend)\end{Bmatrix}$
	\end{algorithmic}
\end{algorithm}

\subsection{Fusing the Estimators}

We designed a Fusioner based on a multi-layer perceptron~\cite{hampshire1991equivalence} to output the final triples trustworthiness values. A simple way to combine the above Estimators is to splice their outputs into a feature vector $f(s)$ for each triple $s=(h,r,t)$ and,
\begin{equation}f(s)=\begin{bmatrix}RR(h,t), p(E(s)), RP(s) \end{bmatrix}\end{equation}
The vector $f(s)$ will be inputted into the Fusioner and transformed passing multiple hidden layers. The output layer is a binary classifier by assigning a label of $y$ = 1 to true tuples and a label of $y$ = 0 to fake ones. A nonlinear activation function (logistic sigmoid) is used to calculate $p(y = 1|f(s))$ as,
\begin{equation}
\left\{\begin{matrix}
h_{i}=\sigma (W_{h_{i}}f(s) + b_{h_{i}}) \\ 
p(y=1\mid f(s))=\varphi (W_{o}h + b_{o})
\end{matrix}\right.
\end{equation}
Where $h_{i}$ is the $i_{th}$ hidden layer, $W_{h_{i}}$ and $b_{h_{i}}$ are the parameter matrices to be learned in the $i_{th}$ hidden layer, and $W_{o}$ and  $b_{o}$ are the parameter matrices of the output layer.

\section{Experiments}
\label{sec:experiments}

\subsection{Experimental Settings}

We focus on Freebase~\cite{Bollacker2008}, which is one of the most popular real-world large-scale knowledge graphs, and we perform our experiments on the FB15K~\cite{Bordes2013}, which is a typical benchmark knowledge graph extracted from Freebase. As for the FB15K, there are 1,345 relations and 14,951 entities and the corresponding 592,213 triples. We use all of the 592,213 triples to construct graphs which are described in Section 3.1. Each head entity is the core of a graph, so we can construct 14,951 graphs. These graphs are used for ResourceRank algorithm and Reachable paths inference algorithm.

There are no explicit labelled errors in the FB15K. Considering the experience that most errors in real-world KGs derive from the misunderstanding between similar entities, we use the methods described in~\cite{Xie2017a} to generate fake triples as negative examples automatically where the picked entity should once appear at the same position. For example, (Newton, Nationality, American) is a potential negative example rather than the obvious irrational (Newton, Nationality, Google), given a true triple (Newton, Nationality, England), as England and American are more common as the tails of Nationality. 
We assure that the number of negative examples is equal to that of positive examples. In a random but quantitatively balanced manner, one of the three kinds of fake triples may be constructed for each true triple: one by replacing head entity, one by replacing relationship, and one by replacing tail entity. We assign a label of 1 to positive examples and 0 to negative examples. Therefore, we build a corpus for our experiments which contains double 592,213 triples. It is separated for training (double 483,142 triples), valid (double 50,000 triples), and testing (double 59,071 triples)

We implement the neural network using the Keras library\footnote{https://github.com/keras-team/keras}\footnote{The code can be obtained from https://github.com/TJUNLP/TTMF.}. The dimension of the entity and relation embeddings is 100. The batch size is fixed to 50. We use early stopping~\cite{graves2013speech} based on the performance on the validation set. The number of RNN units is 100. Parameter optimization is performed with the Adam optimizer~\cite{kingma2014}, and the initial learning rate is 0.001. In addition, to mitigate over-fitting, we apply the dropout method~\cite{srivastava2014} to regularize our model.

In addition, there are some adjustable parameters during the model training.
$\theta$ is the probability that the resource flow from a node directly jumps to a random node. According to the value in the PageRank algorithm~\cite{Page1998,Broder2000}, we set $\theta=0.15$. We set $K=4$ and $TopK=3$. If the two parameters are set too large, the consumption of model training will be greatly increased. If they are set too small, the related algorithm will be affected. Thus this is the trade-off after repeated attempts. The relation-specific threshold $\delta _{r}$ can be searched via maximizing the classification accuracy on the validation triples, which belong to the relation $r$.

\begin{table*}
	\caption{Evaluation results on the Knowledge graph error detection.}
	\centering
	\setlength{\tabcolsep}{3pt}
	\begin{tabular}{p{50pt}|p{30pt}p{30pt}p{30pt}p{30pt}p{30pt}p{30pt}p{30pt}p{50pt}p{55pt}p{50pt}}
		\toprule[1pt]
		Models& MLP	&Bilinear &TransE &TransH &TransD &TransR &PTransE &Ours\_TransE& Ours\_PTransE		&Ours\_TransH \\\hline
		Accuracy&	0.833	&0.861	&0.868	&	0.912	&0.913	&0.902	&0.941	&0.977	&0.978	&\textbf{0.981}\\
		F1-score	& 0.846&0.869&	0.876&	0.913&	0.913&	0.904&	0.942&	0.975&	0.979&	\textbf{0.982}\\	\bottomrule[1pt]
	\end{tabular}
	\label{tab2}
\end{table*}

\subsection{Interpreting the Validity of the Trustworth- iness}

\begin{figure}
	\centering
	\includegraphics[width=0.485\textwidth]{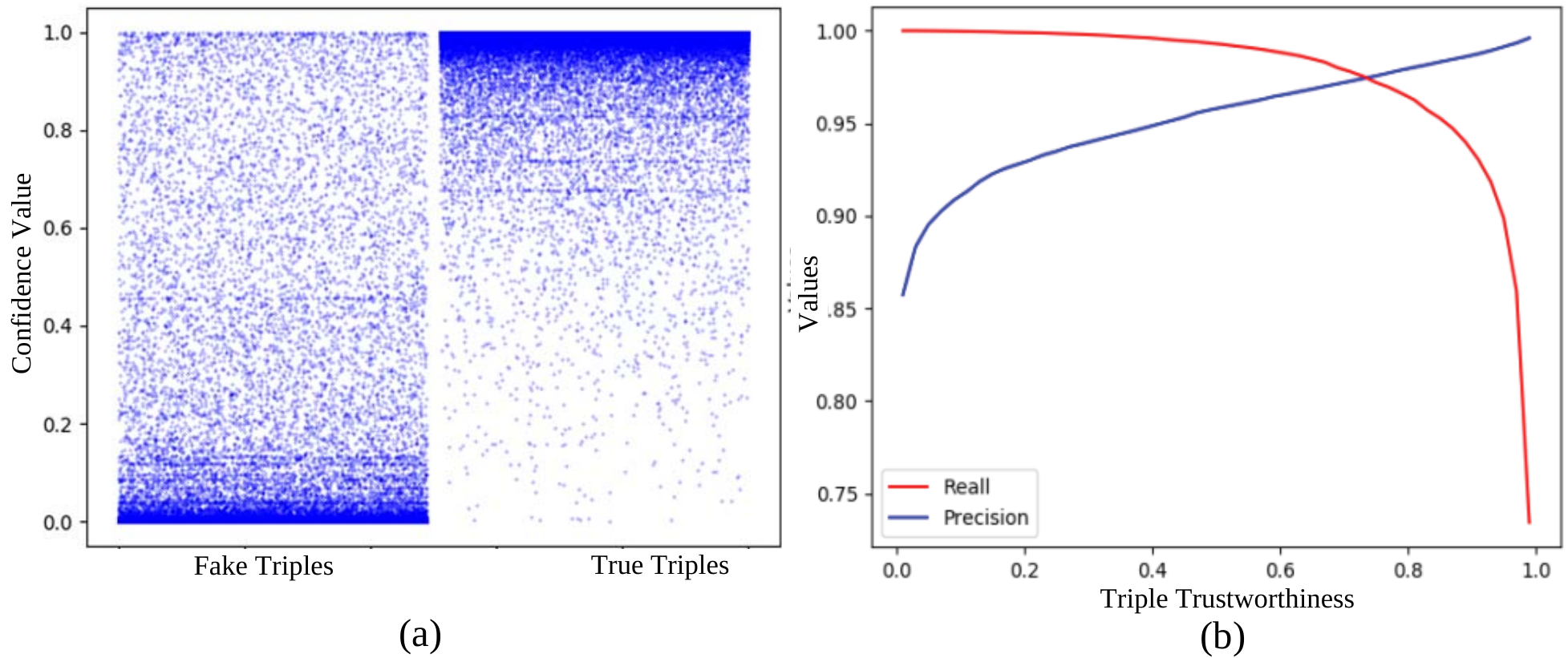}
	\caption{(a) The scatter plot of the triple confidence values distribution. (b) The various value cures of precision and recall with the triple confidence values.}
	\label{fig6}
\end{figure}

To verify whether the triple trustworthiness outputs from KGTtm is valid, we do the following analysis on the test set.

We display the triple confidence values in a centralized coordinate system, as shown in figure~\ref{fig6}(a). The left area shows the distribution of the values of the negative examples, while the right area shows that of the positive examples. It can be seen that the confidence values of the positive examples are mainly concentrated in the upper region (> 0.5)\footnote{The output-layer of our model is a binary classifier, and the confidence value of a triple is the probability that the triple is predicted as label=1, therefore we choose a threshold of 0.5.}. In contrast, the values of the negative examples are mainly concentrated in the lower region (< 0.5). It is consistent with the natural law of judging triple trustworthiness, proving that the triple confidence values output from our model are meaningful.

%\begin{figure}
%	\centering
%	\includegraphics[width=0.25\textwidth]{figure/jia6.pdf}
%	\caption{The scatter plot of the triple confidence values distribution.}
%	\label{fig6}
%\end{figure}

%\begin{figure}
%	\centering
%	\includegraphics[width=0.25\textwidth]{figure/jia7.pdf}
%	\caption{The various value cures of precision and recall with the triple confidence values.}
%	\label{fig7}
%\end{figure}

%\begin{figure}
%	\centering
%	\subfigure[]{
%		\label{fig6:a} %% label for first subfigure
%		\includegraphics[width=0.25\textwidth]{figure/jia6.pdf}}
%	\hspace{1in}
%	\subfigure[]{
%		\label{fig6:b} %% label for second subfigure
%		\includegraphics[width=0.25\textwidth]{figure/jia7.pdf}}
%	\caption{(a) The scatter plot of the triple confidence values distribution. (b) The various value cures of precision and recall with the triple confidence values.}
%	\label{fig6} %% label for entire figure
%\end{figure}

In addition, by dynamically setting the threshold for the triple confidence values (only if the value of a triple is higher than the threshold can it be considered trustworthy.), we can measure the curves of the precision and recall of the output, as shown in figure~\ref{fig6}(b). As the threshold increases, the precision continues to increase, and the recall continues to decrease. When the threshold is adjusted within the interval [0, 0.5], there is no obvious change in the recall, and it remains at a high level. However, if the threshold is adjusted within the interval [0.5, 1] , the recall tends to decline. In particular, the closer the threshold is to 1, the greater the decline rate will be. These show that the positive examples universally have higher confidence values (> 0.5). Moreover, the precision has remained at a relatively high level, even when the threshold is set to a small value, which indicates that our model can identify the negative instances well and assign them a small confidence value.

\begin{table}
	\caption{Evaluation results of the three type noises.}
	\label{table3}
	\setlength{\tabcolsep}{3pt}
	\begin{tabular}{p{35pt}|p{28pt}p{28pt}|p{28pt}p{28pt}|p{28pt}p{28pt}}
		\toprule[1pt] 
		Models &\multicolumn{2}{|c|}{$(h, r, ?)$} &\multicolumn{2}{|c|}{$(h, ?, t)$} &\multicolumn{2}{|c}{$(?, r, t)$} 
		\\\cline{2-7}
		&Recall	&Quality&	Recall	&Quality&	Recall	&Quality \\\hline
		MLP	 &0.970	 &0.791	 &0.912	 &0.735	 &0.978	 &0.844 \\
		Bilinear&	0.936&	0.828&	0.904&	0.807&	0.973&	0.907\\
		TransE&	0.960&	0.796	&0.927&	0.759&	0.959&	0.786\\
		TransH&	0.935&	0.826&	0.927&	0.811&	0.955&	0.850\\
		TransD&	0.942	&0.838&	0.909&	0.804&	0.954&	0.853\\
		TransR&	0.964&	0.872&	0.921&	0.829&	0.972&	0.868\\
		PTransE&	0.944	&0.841	&0.973	&0.888&	0.957	&0.863\\
		Ours&	\textbf{0.987}&	\textbf{0.943}&	\textbf{0.977}&	\textbf{0.923}&	\textbf{0.994}&	\textbf{0.959}
		\\	\bottomrule[1pt]
		
	\end{tabular}
	\label{tab3}
\end{table}%

\begin{table}
	\caption{Evaluation results of each single estimator on the Knowledge graph error detection.}
	\centering
	\setlength{\tabcolsep}{3pt}
	\begin{tabular}{p{40pt}|p{50pt}p{60pt}p{25pt}p{35pt}} 
		\toprule[1pt] 
		Models &TEF(TransE)&ResourceRank&RPI& KGTtm \\\hline	
		Accuracy&	0.868 &	0.811 &	0.881  &	0.977 \\ 	\bottomrule[1pt] 
	\end{tabular}
	\label{tab4}
\end{table}%

\subsection{Comparing With Other Models on The Knowledge Graph Error Detection Task}

The Knowledge graph error detection task is to detect possible errors in the KG according to their triple trustworthy scores. Exactly, it aims to predict whether a triple is correct or not, which could be viewed as a triple classification task~\cite{socher2013reasoning}. 

We give several evaluation results. (1) The accuracy of classification. The decision strategy for classification is that if the confidence value of a testing triple $(h, r, t)$ is below the threshold 0.5, it is predicted as negative, otherwise, it is positive. (2) The maximum F1-score when the given threshold is at [0, 1].

As shown in table~\ref{tab2}, our model has better results in terms of accuracy and the F1-score than the other models. 
The Bilinear model~\cite{Li2016}~\cite{Nickel2011}~\cite{Yang2014} and Multi layer perceptron (MLP) model~\cite{Dong2014} ~\cite{Li2016} have been widely applied to the KG related tasks. They can calculate a score for the validity of triples through operations, such as tensor decomposition and nonlinear transformation. Here we convert the scores to the confidence values using the sigmoid function. Compared with the Bilinear and MLP models, our model shows improvements of more than 10\% in the two evaluation indicators.
We use the TEF algorithm (as illustrated in Section 3.2) to transform the output of the embedding-based models of TransE, TransH, TransD, TransR, and PTransE into triple confidence values. These embedding-based models are better than the traditional method, but their results are affected by the quality of the embeddings. In comparison, our model does not rely on word embeddings. We introduce different embeddings into our model, as shown by Ours\_TransE, Ours\_TransH, and Ours\_PTransE, which have very subtle effects. Since our model makes full use of the internal semantic information of the triple and the global inference information of the knowledge graph, it is more robust to achieve the three-level measure of trustworthiness.

\subsection{Analyzing the ability of models to tackle the three type noises.}

Three types of errors or noises are generated by replacing the head entity, tail entity, or relation in the triples. We measure the ability of the models to recall positive cases from candidate triples doped with a large number of noises. 

We select only true triples (positive examples) in test set and divide them into three categories: all pairs of head and relation $(h, r, ?)$, all pairs of head and tail $(h, ? , t)$, and all pairs of tail and relation $(?, r, t)$. Then  complement all empty positions with the objects in entity set or relationship set. In this way, for a certain pair of head and relation or a pair of tail and relation, 14,951 candidate triples can be constructed respectively. Similarly, for a pair of head and tail entities, 1,345 candidate triples can be generated. As for a complemented triple $(h, r, t)$, we calculate its confidence value. When the value is higher than the threshold (> 0.5), we judge it to be correct.
Two evaluation metrics are conducted as: (1) The recall of true triples in the test set (Recall). (2) The average trustworthiness values across each set of true triples (Quality)~\cite{Li2016}.

By analyzing the results in table~\ref{tab3}, we find that our model achieves a higher recall on the three types of test sets compared to other models, it shows that our model can more accurately find the right from noisy triples. The average trustworthiness values of our model is higher than that of others, which show that our model can better identify the correct instances and with high confidence values. In addition, our model achieves the best results on the $(?, r, t)$ set, but the worst among the $(h, ?, t)$ set. It can be found that the output of almost all models satisfies this phenomenon. It is difficult to judge the relation types of an entity pair, because there may be various relations between a certain entity pair, which increases the difficulty of model judgment.

\subsection{Analyzing the Effects of Single Estimators}

To measure the effect of single Estimators, we separate each Estimator as an independent model to calculate the confidence values for triples. The results in the knowledge graph error detection are shown in table~\ref{tab4}. It can be found that the accuracy obtained by each model is above 0.8, which proves the effectiveness of each Estimator. Among them, the Reachable paths inference (RPI) based method achieves better results than the other two Estimators. After combining all the Estimators, the accuracy obtained by the global model (KGTtm) has been greatly improved, which shows that our model has good flexibility and scalability. It can well integrate multiple aspects of information to obtain a more reasonable trustworthiness.

It is worth emphasizing that our model is flexible and easy to extend. The newly added estimators can train their parameters together with the model frame. In addition, the confidence value generated by a single estimator can be extended to the feature vector $f(s)$ straightly.

\section{Conclusion}

In this paper, to eliminate the deviation caused by the errors in the KG to the knowledge-driven learning tasks or applications, we establish a knowledge graph triple trustworthiness measurement model (KGTtm) to detect and eliminate errors in the KG. The KGTtm is a crisscrossing neural network structure, it evaluates the trustworthiness of the triples from three perspectives and synthetically uses the triple semantic information and the global inference information of the knowledge graph. Experiments were conducted on the popular knowledge graph Freebase. The experimental results confirmed the capabilities of our model. In the future, we will explore adding more estimators to the model to further improve the effectiveness of the trustworthiness. We will also try to apply the trustworthiness to more knowledge-based applications.

\begin{acks}

  The authors would also like to thank the anonymous referees for
  their valuable comments and helpful suggestions. 
  This work was supported in part by the National Natural Science Foundation of China under Grant 71571136, and in part by the Project of Science and Technology Commission of Shanghai Municipality under Grant 16JC1403000 and Grant 14511108002.

\end{acks}

\bibliographystyle{ACM-Reference-Format}
\balance 
\bibliography{TTMF}

\end{document}